\pdfoutput=1
\documentclass[11pt]{article}

\usepackage[utf8]{inputenc}
\usepackage[T1]{fontenc}
\usepackage{lmodern}
\usepackage[a4paper,margin=1in]{geometry}
\usepackage{microtype}
\usepackage{amsmath,amssymb}
\usepackage{bm}
\usepackage{graphicx}
\usepackage{booktabs}
\usepackage{tabularx}
\usepackage{array}
\usepackage{makecell}
\usepackage[font=small,labelfont=bf]{caption}
\usepackage{threeparttable}
\usepackage[authoryear,round]{natbib}
\usepackage{authblk}
\usepackage{xcolor}
\usepackage{xurl}
\usepackage[colorlinks=true,linkcolor=blue!60!black,citecolor=blue!60!black,urlcolor=blue!60!black]{hyperref}

\graphicspath{{figures/}}

\hypersetup{
  pdftitle={Lens: Training-Free Semantic Search for Cross-Tile Object and Scene Retrieval in Satellite Imagery},
  pdfauthor={Ata Akbari Asanjan, Anit Upadhyaya, Jeremy Kravitz, Subash Yeggina},
  pdfsubject={Training-free semantic search and retrieval in satellite imagery},
  pdfkeywords={content-based image retrieval, semantic search, training-free, satellite imagery, remote sensing, Sentinel-2, self-supervised learning, vision foundation models, DINOv3, feature debiasing, feature upsampling, AnyUp, Earth observation}
}

\newcommand{\R}{\mathbb{R}}
\providecommand{\doi}[1]{\href{https://doi.org/#1}{doi:\nolinkurl{#1}}}
\newcommand{\vect}[1]{\bm{#1}}
\DeclareMathOperator{\normalize}{normalize}

\newcolumntype{L}[1]{>{\raggedright\arraybackslash}p{#1}}

\title{\textbf{Lens: Training-Free Semantic Search for Cross-Tile Object and Scene Retrieval in Satellite Imagery}}

\author[1]{Ata Akbari Asanjan\thanks{Corresponding author: \href{mailto:ata.akbari@pixxel.space}{ata.akbari@pixxel.space}}}
\author[2]{Anit Upadhyaya}
\author[1]{Jeremy Kravitz}
\author[2]{Subash Yeggina}
\affil[1]{Pixxel Space Technologies, El Segundo, United States}
\affil[2]{Pixxel Space India Pvt Ltd, Bengaluru, India}

\date{}

\begin{document}

\maketitle

\begin{abstract}
We developed Lens, a training-free system for semantic object and scene search across large satellite mosaics that requires no detector trained for the class of interest. Lens represents each tile with a frozen, self-supervised vision transformer (DINOv3), removes the positional bias in its dense features by projecting them onto the subspace orthogonal to a positional basis estimated once from a constant image, and centers the debiased features on the mean feature computed across the whole area before comparison. Lens forms a query by averaging the centered features of a small neighborhood of patches around a chosen point, or of an entire tile for scene-level search, and matches it against the index by cosine similarity. It stores the centered patch features in a byte-quantized similarity index that supports real-time range search over a mosaic of thousands of patches, and a coarser whole-tile embedding index for scene-level retrieval. A guided upsampling network (AnyUp) runs once per tile to recover pixel-accurate boundaries beyond the native feature grid, and the result is cached so later queries reuse it. We evaluated Lens on the 5,128-tile DFC2020 test set and ran it as an interactive tool on five Sentinel-2 areas spanning agricultural, industrial, and desert terrain; the pipeline is not tied to a specific sensor. On DFC2020 a single click reaches 0.09 to 0.80 pixel F1 across eight land-cover classes at the per-class best threshold, with a mean F1 of 0.39, and it beats a pairwise INSID3 baseline on six of the eight classes at roughly four hundred times lower query cost. Indexing 1.31 million patches takes 169~s on one CPU thread, and a click returns a rendered overlay in 6 to 10~ms.

\end{abstract}

% ===========================================================================
\section{Introduction}
\label{sec:intro}

Satellite archives have grown faster than our ability to ask questions of them. Copernicus alone delivers global Sentinel-2 coverage every five days at 10~m \citep{drusch2012}. The bottleneck is no longer acquisition. It is that almost every question an analyst wants to ask has to be converted into a supervised learning problem first. An analyst who wants to know how many center-pivot irrigation fields are active this summer in the Wadi As-Sirhan basin has to collect training examples, annotate them, train a detector, tune it, and run it over the area: weeks of work, most of it annotation, and none of it transferring if the next question is about solar arrays in Rajasthan. Detectors get built for the few classes that justify the investment, and the long tail goes unexamined.

Lens changes the economics of that first query. It is a retrieval system rather than a detector: the analyst points at one example and the system returns everything in the area that resembles it. No labels are collected, no model is trained, and the class is whatever the analyst just clicked on. Retrieval by example is an old idea with a substantial remote sensing literature behind it \citep{sumbul2020}, and what makes it newly practical is the quality of self-supervised features. DINOv3 \citep{simeoni2025} produces dense patch descriptors that are semantically meaningful without task supervision and holds up on aerial imagery, so in principle one can index those descriptors and search them directly.

In practice this does not work, and understanding why is the substance of this paper. Two distortions sit on top of the semantic signal. Dense DINOv3 features carry a component that depends on where a patch sits in the image rather than what it contains \citep{cuttano2026}. Separately, when every tile in an area shows the same terrain, the descriptors cluster so tightly that cosine similarity loses its range. Both are removable with linear operations that cost nothing at query time, but they are not equally important: the positional projection carries most of the benefit, and area centering adds to it only when the target occupies a small fraction of the area. We report that asymmetry rather than the cleaner story we expected. The system is measured on the DFC2020 test set and exercised as a deployed tool on five Sentinel-2 areas.

% ===========================================================================
\section{Problem Statement}
\label{sec:problem}

Given a mosaic covering an area of interest, divided into tiles, and a single point an analyst has clicked, return every location in the mosaic whose local appearance resembles the neighborhood of that point, with the resemblance controllable by one scalar the analyst can adjust.

Five constraints make this harder than it sounds. The system must not train, because any step that fits parameters to the current area reintroduces the delay we are trying to eliminate. Similarity must be comparable across tiles, which fails if descriptors encode position, since patches at the same offset then match each other regardless of content. One threshold must work for the whole area, which requires the similarity distribution to have real spread. Response must be interactive on a laptop CPU. And the output must be sharper than the feature grid: a ViT with 16-pixel patches on a 512-pixel tile yields a $32 \times 32$ grid, so on 10~m imagery each descriptor covers 160~m and a thresholded query returns a blocky mask whose edges bear little relation to the object boundary.

% ===========================================================================
\section{Objectives}
\label{sec:objectives}

The work had four objectives: to build a retrieval pipeline over satellite mosaics using only frozen, publicly available models; to identify and correct the distortions that prevent frozen dense features from supporting cross-tile search, and establish that those corrections are cheap at index time and free at query time; to make the similarity scale interpretable enough that one threshold behaves consistently across a heterogeneous area; and to reach interactive latency on commodity hardware through quantized indexing and query-independent caching.

% ===========================================================================
\section{Literature Review}
\label{sec:related}

\subsection{Retrieval in remote sensing archives}

Content-based retrieval of remote sensing imagery has been studied for two decades, moving from handcrafted spectral and textural descriptors to learned representations trained on labeled archives \citep{sumbul2020}. Two properties of that work matter here. The retrieval unit is almost always a whole image, so these systems answer ``which tiles look like this tile'' rather than ``where in the mosaic is this object'', and the representations are learned from labeled archives, which ties their vocabulary to the label set. Lens retrieves at patch level within a mosaic, and its representation is never trained on remote sensing labels at all.

\subsection{Foundation models and embedding fields for Earth observation}

The response to label dependence has largely been to pretrain domain-specific foundation models. SatMAE \citep{cong2022} adapts masked autoencoding to temporal and multispectral data, and Scale-MAE \citep{reed2023} conditions positional encoding on ground sample distance. Prithvi-EO-2.0 \citep{szwarcman2025} trains on 4.2 million multitemporal samples with explicit temporal and location embeddings, and TerraMind \citep{jakubik2025} is pretrained on nine geospatial modalities. RemoteCLIP \citep{liu2024} instead aligns imagery with text so retrieval can be driven by a written query. What most of these share is that the pretrained encoder is a starting point: reaching a usable output for a specific class still means attaching a head and training it. RemoteCLIP is the exception, but the queries we care about are often ones the analyst cannot name well, and a click expresses them where a noun phrase does not.

A newer line of work targets something closer to what Lens needs, a dense field of reusable descriptors already covering the ground. AlphaEarth Foundations \citep{brown2025} produces a 64-dimensional embedding for every 10~m cell for every year, quantized to 8 bits, and TESSERA \citep{feng2026} encodes Sentinel-1 and Sentinel-2 time series into 128-dimensional per-pixel embeddings at the same resolution. Both converge on two conclusions Lens arrives at independently: descriptors should be reusable without retraining, and they should be byte-quantized, because storage dominates at scale. Why not index one of these fields directly? Because they summarize each cell's spectral and temporal trajectory, which is what per-pixel classification needs. A center-pivot circle is not a property of any one pixel; it is a spatial arrangement of many, and recognizing it from a single click needs local geometry and texture, which a ViT patch descriptor carries and a per-pixel temporal embedding does not. Concatenating the two is an experiment we have not run.

\subsection{Self-supervised dense features and their artifacts}

DINO \citep{caron2021} showed that a vision transformer trained by self-distillation develops features that segment objects with no segmentation supervision, and DINOv2 \citep{oquab2023} turned that into a general-purpose backbone whose frozen features matched or beat weakly supervised alternatives on dense tasks. DINOv3 \citep{simeoni2025} scales the recipe and adds Gram anchoring, which stops dense features degrading over long schedules. That last property is what Lens rests on, since an index is only worth building if the descriptors going into it are good enough to use frozen.

The descriptors are not clean. \citet{cuttano2026} showed that DINOv3 dense features carry a systematic positional bias, so patches at the same image coordinates produce spurious matches across images. Their INSID3 system removes it by estimating the positional subspace from images with no semantic content and projecting features onto its orthogonal complement, reporting gains of 1.4 to 6.6 PCK points on SPair-71k from the correction alone. Lens borrows that projection for a different purpose: INSID3 matches one reference image against one target image at a time, whereas we build a persistent index over an entire corpus and search it repeatedly. The correction transfers because the positional basis is estimated from a constant image and is therefore independent of any particular tile, which is what a shared index requires.

\subsection{Prompt-driven segmentation}

The Segment Anything Model \citep{kirillov2023} segments from point or box prompts, and the interaction pattern resembles ours. SAM~2 \citep{ravi2025} generalizes it to video. For both, the mask carries no semantic label, so neither can say where else the same thing appears, and transfer to overhead imagery is uneven: \citet{ren2024} found competitive zero-shot results on some overhead tasks but clearly worse than supervised models on most.

SAM~3 \citep{carion2026} changes this picture and is the closest work to ours. Its promptable concept segmentation takes a short noun phrase or a set of image exemplars and returns masks for every matching instance, so the step SAM leaves to the user is now inside the model. Two architectural differences remain. SAM~3 answers a query by running an 848\,M-parameter detector over one image, whereas Lens looks up an index built once for the whole area, so every later query, including each drag of the threshold slider, costs only a search. And SAM~3 returns instances, where Lens returns a continuous similarity field. A comparison against SAM~3 with exemplar prompts is the most informative experiment this paper does not contain.

\subsection{Feature upsampling}

FeatUp \citep{fu2024} restores spatial resolution using multiview consistency, but both its variants must be trained per encoder. AnyUp \citep{wimmer2026} removes that restriction with a feature-agnostic layer, so a single model trained once on DINOv2 ViT-S features upsamples features from encoders it has never seen, including DINOv3. Lens uses AnyUp for exactly this reason: we need pixel-accurate boundaries without training an upsampler for our configuration.

% ===========================================================================
\section{Methodology}
\label{sec:method}

\subsection{Overview}

\begin{figure}[t]
  \centering
  \includegraphics[width=\textwidth]{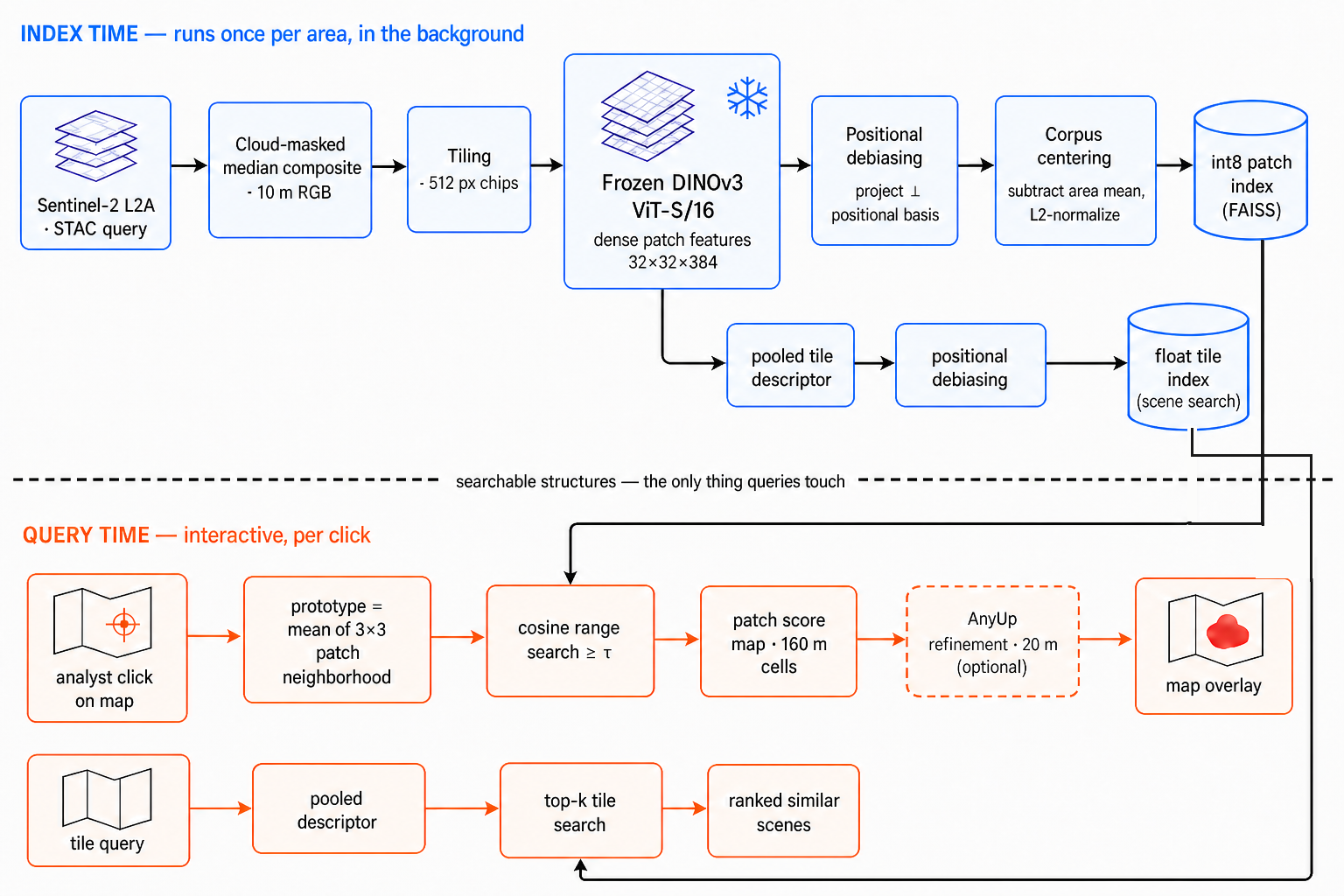}
  \caption{Architecture of the Lens indexing and query pipeline.}
  \label{fig:pipeline}
\end{figure}

Lens has two phases (Figure~\ref{fig:pipeline}). Indexing runs once per area of interest, in the background: it takes a bounding box and a date range, retrieves Sentinel-2 scenes, builds a cloud-free composite, divides it into tiles, encodes each tile into dense descriptors, applies the two corrections, and writes the results into two indices. Querying runs on every click and touches only those structures, forming a descriptor from the click, range-searching the patch index, and painting the hits back onto the map, optionally refined to pixel resolution.

\subsection{Evaluation corpus and study areas}

Accuracy is measured on the DFC2020 test set, the land-cover benchmark from the 2020 IEEE GRSS Data Fusion Contest \citep{robinson2021}. It contains 5,128 Sentinel-2 tiles of $256 \times 256$ pixels at 10~m, each with a dense label map in a simplified ten-class IGBP scheme. We use B04, B03, and B02 only, matching the three-band constraint of the pipeline, and obtain the tiles from the Copernicus-Bench mirror, which repackages the original imagery under CC BY 4.0. Two properties make it the right primary reference: the labels are per-pixel, so a result can be scored where it actually falls, and the tiles span many biomes, so a query has to generalize across a corpus no single background dominates. The mirror strips georeferencing, so we treat the split as one corpus of 1,312,768 patches with one shared mean, the least favorable configuration for area centering.

The deployed tool is exercised separately on the five areas in Table~\ref{tab:sites}, which differ in terrain, object type, and density. Al Jawf is the primary development site, with center-pivot irrigation on uniform desert sand, the hardest case for background separation and the easiest for visual verification. Imperial Valley has the same object class at higher density against a more varied background, Bhadla and Atacama contain photovoltaic arrays, and Rotterdam is the outlier, with dense port infrastructure at many scales. These five carry the timing results and a secondary check against ESA WorldCover, not the headline accuracy numbers.

\begin{table}[t]
  \centering
  \small
  \caption{Sentinel-2 areas of interest used for the deployed-tool experiments.}
  \label{tab:sites}
  \begin{threeparttable}
  \setlength{\tabcolsep}{4pt}
  \begin{tabularx}{\textwidth}{@{}l l l l c X@{}}
    \toprule
    \thead[l]{Site} & \thead[l]{Location} & \thead[l]{Bounding box\\(W, S, E, N)} & \thead[l]{Date window} & \thead{Scenes} & \thead[l]{Dominant content} \\
    \midrule
    Al Jawf         & Saudi Arabia    & \makecell[tl]{38.15, 30.40,\\38.30, 30.52}             & \makecell[tl]{2024-06-01 to\\2024-09-01} & 38 & Center-pivot irrigation on desert \\
    Bhadla          & India           & \makecell[tl]{71.88, 27.48,\\72.02, 27.58}             & \makecell[tl]{2024-01-01 to\\2024-04-01} & 21 & Photovoltaic arrays, Thar Desert \\
    Imperial Valley & \makecell[tl]{California,\\USA} & \makecell[tl]{$-115.62$, 32.95,\\$-115.48$, 33.07} & \makecell[tl]{2024-05-01 to\\2024-08-01} & 14 & Dense center-pivot agriculture \\
    Rotterdam       & Netherlands     & \makecell[tl]{4.05, 51.88,\\4.22, 51.98}               & \makecell[tl]{2024-05-01 to\\2024-08-01} & 8  & Port berths, tanks, industry \\
    Atacama         & Chile           & \makecell[tl]{$-69.52$, $-23.12$,\\$-69.38$, $-22.98$} & \makecell[tl]{2024-01-01 to\\2024-04-01} & 35 & Solar arrays, high desert \\
    \bottomrule
  \end{tabularx}
  \begin{tablenotes}\footnotesize
    \item Each site is nine 512-pixel tiles except Atacama, which is twelve.
  \end{tablenotes}
  \end{threeparttable}
\end{table}

Imagery is Sentinel-2 Level 2A surface reflectance, queried through a SpatioTemporal Asset Catalog interface and cloud-masked using the scene classification layer. We take the per-pixel median over the date window, which absorbs the cloud edges the classification layer misses, project to the area's native UTM zone at 10~m while retaining the affine transform, and convert reflectance to 8-bit color by a 2nd to 98th percentile stretch. Tiling uses 512-pixel windows at a stride of 512, with the last window in each row and column snapped flush to the composite edge rather than padded.

\subsection{Dense representation}

Each tile is resized to $512 \times 512$, scaled to the unit interval, normalized with ImageNet channel statistics, and passed through a frozen DINOv3 ViT-S/16. We take the mean of the token sequences from the last four transformer blocks and apply the final layer normalization, following the configuration INSID3 uses; fusing several blocks rather than taking only the last retains more spatial detail while staying semantically organized. We use the natural-image checkpoint rather than the satellite-pretrained one, having found no advantage for the latter on dense separation in our tiles, though that is an observation on a few areas rather than a controlled comparison.

The dense path keeps the rotary position embedding active through the transformer blocks. This may seem contradictory given that we then remove positional information, but dropping the embedding degrades the features rather than cleaning them, because the model was trained with it. The right order is to let the model run as trained and remove the positional component afterwards in feature space.

\subsection{Removing the positional component}
\label{sec:debias}

Two patches at the same grid coordinates in different tiles share a component that has nothing to do with what they show. Within a single image this is tolerable; across an index spanning many tiles it is not, because it creates a background of spurious matches organized by position.

Following \citet{cuttano2026}, we isolate this component by encoding an image with no content, which yields a descriptor map in which every remaining variation is positional by construction. Writing the L2-normalized descriptors of that map as a matrix $\vect{E} \in \R^{C \times N}$ with $C = 384$ channels and $N = 1024$ locations, we subtract the spatial mean of each channel and take the thin singular value decomposition $\vect{E} = \vect{U}\vect{\Sigma}\vect{V}^{\top}$. The first $k = 64$ left singular vectors form an orthonormal basis $\vect{B} \in \R^{C \times k}$ for the positional subspace, and every patch descriptor $\vect{x}$ is projected onto its orthogonal complement and renormalized:
\begin{equation}
  \vect{x}_{\mathrm{deb}} = \normalize\!\left( (\vect{I} - \vect{B}\vect{B}^{\top})\, \vect{x} \right),
  \qquad \normalize(\vect{v}) = \vect{v} / \lVert \vect{v} \rVert_2 .
  \label{eq:debias}
\end{equation}

The basis depends only on the encoder and the input resolution, so it is computed once and cached. That is what makes a shared index sound: every tile in every area is corrected against the same basis, so the corrected descriptors live in a common space. The same projection is applied to the pooled tile descriptor for the scene index, for consistency rather than for measured benefit.

\subsection{Centering on the area mean}
\label{sec:centering}

The second correction addresses a problem the projection does not touch. Consider Al Jawf, where almost every patch is sand. The debiased descriptors encode the content faithfully, but the content is nearly identical everywhere, so they occupy a small cone in feature space and similarity to any query sits in a narrow positive band. The effect is not confined to deserts: on DFC2020 the median score of target patches exceeds the median of everything else by only 0.10. A slider operating on that distribution spends most of its travel doing nothing.

The fix is to remove the shared component explicitly. After every tile has been encoded and debiased, we compute the mean descriptor $\vect{\mu}$ over all patches, store it with the index, and center and renormalize every descriptor and query before comparison:
\begin{equation}
  \vect{z} = \normalize\!\left( \vect{x}_{\mathrm{deb}} - \vect{\mu} \right).
  \label{eq:center}
\end{equation}

Cosine similarity between centered descriptors then measures agreement in departure from the area average rather than in absolute appearance. Sand, being the average, has small residual magnitude in inconsistent directions, while irrigation circles depart from the average consistently and so agree strongly with one another.

\begin{figure}[t]
  \centering
  \includegraphics[width=0.85\textwidth]{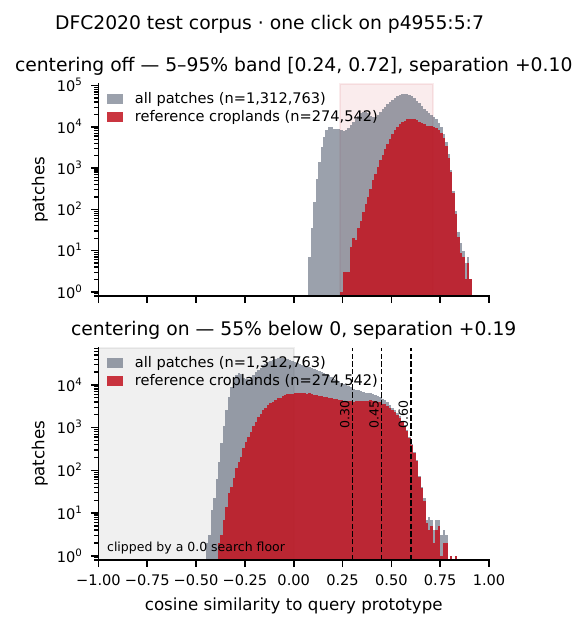}
  \caption{Patch-to-query cosine over the DFC2020 corpus for one protocol click, with and without centering, target patches shaded; the five patches of the clicked prototype are excluded. Median separation rises from 0.10 to 0.19.}
  \label{fig:centering}
\end{figure}

Three details matter for reproducing this. The mean is computed over the entire area, not per tile, since a per-tile mean would make scores incomparable across tiles. It is computed after debiasing, because the positional component is not part of what we want to measure deviation from. And it is persisted alongside the index and applied identically to queries. One consequence has to be designed for: centering moves 55\% of the DFC2020 scores below zero, so the search floor must go negative (Figure~\ref{fig:centering}).

\subsection{Index construction}

Centered descriptors go into a Faiss index \citep{douze2024} configured for inner product over unit vectors, with 8-bit scalar quantization. Vectors carry identifiers encoding their tile and their row and column within that tile's grid, so recovering a hit's map coordinates is arithmetic on the stored affine transform. Byte quantization reduces the index to a quarter of its float size at a negligible cost in fidelity. Measured on the full DFC2020 corpus rather than on synthetic vectors, the quantized index agrees with exact float scoring to within 0.00266 in similarity, returns range sets with a Jaccard overlap of 0.996 against the exact sets, and retains recall at 10 of 0.988. The raw debiased descriptors are kept in memory at float precision, because query formation reads from them and the prototype should be exact. The tile-level index stores one exact float vector per tile and answers top-$k$.

\subsection{Query formation and search}

A click resolves to a tile and to a row and column in that tile's descriptor grid. The prototype is the mean of the centered descriptors in a disk $\mathcal{D}$ of a given radius around that location, renormalized:
\begin{equation}
  \vect{q} = \normalize\!\left( \frac{1}{|\mathcal{D}|} \sum_{j \in \mathcal{D}} \normalize\!\left( \vect{x}_{\mathrm{deb},j} - \vect{\mu} \right) \right).
  \label{eq:query}
\end{equation}

Each descriptor is centered and normalized before averaging, matching what is stored in the index, and the average is then renormalized. Radius 0 uses the single clicked descriptor; radius 1 adds the four edge-adjacent neighbours, giving a five-descriptor prototype covering roughly 480~m across. Search is a range query returning every indexed descriptor whose inner product with the prototype is at least the threshold, with no size limit, so a common target returns a few hundred descriptors and a rare one returns five without either being padded or truncated to a fixed $k$. Hits are painted into a score map at mosaic resolution, one $16 \times 16$ pixel block per hit, georeferenced through the tile transforms and rendered as an overlay.

\subsection{Pixel-level refinement}

A 16-pixel patch on 10~m imagery covers 160~m, so thresholded hits form a coarse blocky mask. We sharpen it with AnyUp \citep{wimmer2026}, applied once per tile. The natural approach would be to upsample the similarity map, which is one channel and therefore cheap. We instead upsample the full 384-channel descriptor map, using the tile's color image as guidance, to a $256 \times 256$ grid, then L2-normalize each output pixel and quantize to 8-bit integers, so refining any later query on that tile is a dot product against the cached array.

The argument for doing the expensive thing is correctness, not cost. Measured on one tile, 384 channels take 1.503~s against 1.141~s for one channel, so the cheap route is not much cheaper. What settles the choice is that AnyUp is not linear in its input features: upsampling the similarity map and taking the similarity of the upsampled descriptors differ by up to 0.591 in cosine, and the descriptor-space one is the similarity the index defines. Because the upsampled descriptors do not depend on the query, they can be cached, which converts a per-click cost into a one-time per-tile cost. Caches are held in memory under a 16-tile least-recently-used policy at 25.2~MB per tile, and written to disk for saved areas at 10.3~MB per tile.

\begin{figure}[t]
  \centering
  \includegraphics[width=\textwidth]{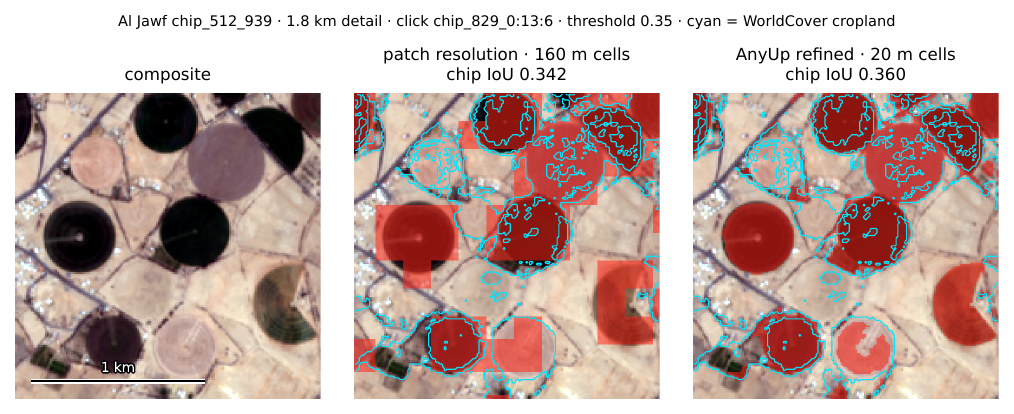}
  \caption{Patch-resolution and refined masks on one Al Jawf tile at threshold 0.35. Intersection over union rises from 0.343 to 0.360 on this tile.}
  \label{fig:refinement}
\end{figure}

\subsection{System implementation}

Lens runs as a client-server application. The backend is a FastAPI service that holds indexed areas in memory, runs indexing as a background job, and exposes endpoints for area creation, overlay rendering, and both query types. The frontend is a React application on MapLibre GL, with a satellite basemap, rectangle drawing, a threshold slider, and a refinement toggle. A saved area reloads with its quantized index, stored mean, raw descriptors, and any cached upsampled tiles. Table~\ref{tab:config} lists the configuration.

\begin{table}[t]
  \centering
  \small
  \caption{Configuration used throughout the experiments.}
  \label{tab:config}
  \begin{threeparttable}
  \begin{tabularx}{0.9\textwidth}{@{}l X@{}}
    \toprule
    \textbf{Component} & \textbf{Setting} \\
    \midrule
    Encoder                  & DINOv3 ViT-S/16, natural-image checkpoint, frozen, 384-dimensional \\
    Tile size                & $512 \times 512$ px at 10~m, giving a $32 \times 32$ descriptor grid \\
    Positional subspace rank & 64 \\
    Patch index              & Faiss, inner product, 8-bit scalar quantization, range search \\
    Tile index               & Faiss, inner product, float, top-$k$ \\
    Refinement               & AnyUp to $256 \times 256$, per-pixel normalized, 8-bit cached \\
    Compute                  & Apple M4, 17.2~GB RAM, single CPU thread \\
    Software                 & Python 3.13.5, PyTorch 2.12.0, Faiss 1.15.0 \\
    \bottomrule
  \end{tabularx}
  \begin{tablenotes}\footnotesize
    \item Every timing was measured on this configuration with OpenMP, MKL, and Faiss pinned to one thread. Megabytes are $10^6$ bytes.
  \end{tablenotes}
  \end{threeparttable}
\end{table}

\subsection{Evaluation protocol}
\label{sec:protocol}

A patch takes the majority class of the labels under its $16 \times 16$ pixel footprint, and a class qualifies if it has at least 200 labelled patches, which admits eight of the ten DFC2020 classes. For each class we simulate 30 clicks on randomly chosen patches whose majority label is the target, under a fixed random seed, and form the same radius-1 prototype the deployed system forms. We then score every patch in the corpus exactly, with a batched float matrix product rather than a range search, because 1.3 million patches make per-seed range results the runtime bottleneck; the shipped int8 path is validated against this exact scoring separately. The five patches of the seed's own prototype disk are excluded so a query cannot be rewarded for finding itself. Thresholds sweep from $-0.20$ to 0.85 in steps of 0.01, and we report precision, recall, and F1 at the 0.30, 0.45, and 0.60 operating points and at the per-class F1-optimal threshold, with average precision, as means over the 30 clicks. Pixel-level metrics are primary, each patch weighted by the labelled positive pixels under it, and all four ablation arms run on the same seeds, so comparisons between them are paired.

Two baselines share this protocol. INSID3 in category mode is identical by construction to our debiased-only ablation, since the features are the same. INSID3 in its native pairwise mode is run separately, one reference tile against one target tile at a time, subsampled to 300 random target tiles per class, and scored generously by giving each pair its own F1-optimal threshold. As a secondary check, the five areas of Table~\ref{tab:sites} are evaluated against ESA WorldCover 2021 v200 \citep{zanaga2022} reprojected onto each mosaic grid. No reference used here was hand-annotated by us.

% ===========================================================================
\section{Results and Discussion}
\label{sec:results}

\subsection{Indexing cost}

Encoding and index construction for a nine-tile area take about 3~s on one CPU thread, and the 1.31 million patches of DFC2020 take 169~s. Scene retrieval and compositing dominate the wall-clock time an analyst waits for a new area, and depend on the catalog rather than on Lens.

The cost that remains splits almost entirely one way. The encoder forward pass accounts for 97.0\% to 99.5\% of index construction across the six corpora, which the arithmetic predicts: one 512-pixel tile costs 64 GFLOPs to encode, while debias, centering, and quantization together cost 107,520 FLOPs per patch, so the corrections add nothing measurable to index time. Memory is the binding constraint instead. For Al Jawf the searchable index is 3.63~MB, while the refinement cache, warm across all nine tiles, is 226.5~MB, sixty times the index it refines, which is why the least-recently-used policy exists (Table~\ref{tab:indexing}).

\begin{table}[t]
  \centering
  \footnotesize
  \setlength{\tabcolsep}{3.5pt}
  \caption{Indexing time and memory. DFC2020 is the primary evaluation corpus; the five sites are the deployed-tool areas of Table~\ref{tab:sites}.}
  \label{tab:indexing}
  \begin{threeparttable}
  \begin{tabular}{@{}l r r r r r r r@{}}
    \toprule
    \thead[l]{Corpus} & \thead[r]{Tiles} & \thead[r]{Patches} & \thead[r]{Index\\time (s)} & \thead[r]{Encoder\\share} & \thead[r]{Patch\\index (MB)} & \thead[r]{Descriptor\\cache (MB)} & \thead[r]{Refinement\\cache,\\warm (MB)} \\
    \midrule
    DFC2020 test split & 5,128 & 1,312,768 & 168.8 & 97.5\% & 514.6 & 1,008.2 & n/a \\
    Al Jawf            & 9     & 9,216     & 3.42  & 97.0\% & 3.63  & 14.2    & 226.5 \\
    Atacama            & 12    & 12,288    & 3.89  & 99.4\% & 4.84  & 18.9    & 302.0 \\
    \bottomrule
  \end{tabular}
  \begin{tablenotes}\footnotesize
    \item Bhadla, Imperial Valley, and Rotterdam match Al Jawf on every column, with index times of 2.92, 3.02, and 3.00~s. Index time excludes scene retrieval and compositing. Descriptor cache is fp16 for DFC2020 and fp32 for the sites.
  \end{tablenotes}
  \end{threeparttable}
\end{table}

\subsection{Retrieval accuracy on DFC2020}

Accuracy varies by more than a factor of eight across the eight classes, from 0.09 pixel F1 on wetlands to 0.80 on water, with a mean of 0.39 and a mean average precision of 0.34 (Table~\ref{tab:accuracy}). That spread is not noise. Water and urban are visually coherent at 160~m and separate cleanly. Wetlands, barren, and shrubland are defined by properties that three visible bands at 10~m do not resolve well, and a click on one instance of them describes the others poorly.

\begin{table}[t]
  \centering
  \small
  \setlength{\tabcolsep}{3.5pt}
  \caption{Retrieval accuracy on the DFC2020 test split for the deployed configuration, with the INSID3 pairwise baseline. Pixel-level, means over 30 simulated clicks per class.}
  \label{tab:accuracy}
  \begin{threeparttable}
  \begin{tabular}{@{}l c c c c c c c c@{}}
    \toprule
    \thead[l]{Class} & \thead{Prevalence} & \thead{Best\\threshold} & \thead{Precision} & \thead{Recall} & \thead{F1} & \thead{AP} & \thead{F1 at\\0.30} & \thead{INSID3\\pairwise F1} \\
    \midrule
    Forest           & 0.263 & 0.04 & 0.437 & 0.756 & 0.535 & 0.496 & 0.226 & 0.475 \\
    Shrubland        & 0.056 & 0.13 & 0.131 & 0.499 & 0.193 & 0.122 & 0.128 & 0.247 \\
    Grassland        & 0.094 & 0.16 & 0.211 & 0.524 & 0.286 & 0.219 & 0.200 & 0.211 \\
    Wetlands         & 0.012 & 0.25 & 0.066 & 0.250 & 0.090 & 0.047 & 0.081 & 0.128 \\
    Croplands        & 0.209 & 0.09 & 0.358 & 0.716 & 0.465 & 0.407 & 0.279 & 0.316 \\
    Urban / built-up & 0.110 & 0.27 & 0.501 & 0.680 & 0.561 & 0.531 & 0.511 & 0.228 \\
    Barren           & 0.022 & 0.26 & 0.146 & 0.244 & 0.155 & 0.097 & 0.146 & 0.132 \\
    Water            & 0.233 & 0.13 & 0.789 & 0.831 & 0.797 & 0.824 & 0.573 & 0.381 \\
    \midrule
    \textbf{Mean}    &       &      & \textbf{0.330} & \textbf{0.563} & \textbf{0.385} & \textbf{0.343} & \textbf{0.268} & \textbf{0.265} \\
    \bottomrule
  \end{tabular}
  \begin{tablenotes}\footnotesize
    \item INSID3 pairwise is given each pair's own F1-optimal threshold, a favorable setting it would not have in deployment.
  \end{tablenotes}
  \end{threeparttable}
\end{table}

The comparison against INSID3 in its native pairwise mode is the clearest evidence that a persistent index is worth building. Lens wins six of the eight classes despite the baseline receiving per-pair oracle thresholds, and the two losses, shrubland and wetlands, are the classes where nothing works well. Anchoring on one reference tile cannot cover the appearance range of a class spread across many biomes. The cost difference runs the same way: a pairwise scan of the corpus costs about 188~s per query, against 0.44~s for a range search over the same 1.31 million patches.

Fixed operating points sit above the per-class optimum, which falls between 0.04 and 0.27, so the 0.30 column understates what the system can do; mean F1 falls further to 0.131 at 0.45 and 0.046 at 0.60. The three thresholds trade recall for precision monotonically, which is the mechanism the interface relies on, but an analyst working a corpus this heterogeneous would operate near 0.1 to 0.3.

\subsection{What each correction contributes}

We expected centering to be the decisive correction and the positional projection to be the enabling prerequisite. The ablation (Table~\ref{tab:ablation}) says the reverse.

\begin{table}[t]
  \centering
  \small
  \caption{Ablation over the two corrections on DFC2020, at the per-class best threshold. Means over the eight classes; ``classes won'' counts classes where an arm has the highest F1 of the four.}
  \label{tab:ablation}
  \begin{tabular}{@{}l c c c@{}}
    \toprule
    \textbf{Configuration} & \textbf{Mean F1} & \textbf{Mean AP} & \textbf{Classes won} \\
    \midrule
    Neither correction          & 0.276 & 0.232 & 0 \\
    Centered only, no debiasing & 0.282 & 0.231 & 0 \\
    Debiased only, no centering & 0.349 & 0.306 & 2 \\
    Debiased and centered       & 0.385 & 0.343 & 6 \\
    \bottomrule
  \end{tabular}
\end{table}

Debiasing on its own recovers 0.073 of the 0.109 total gain in mean F1 and improves every one of the eight classes. Centering on its own recovers 0.006, which is nothing. Applied after debiasing it adds a further 0.036 and wins six classes, so the full pipeline is still best, but the ordering is unambiguous: the positional projection does the work, and centering is a conditional improvement on top of it rather than the correction that makes retrieval possible. Centering also loses outright on forest, 0.535 against 0.580, and ties on grassland, the two classes where the target is close to being the corpus itself.

\begin{figure}[t]
  \centering
  \includegraphics[width=0.85\textwidth]{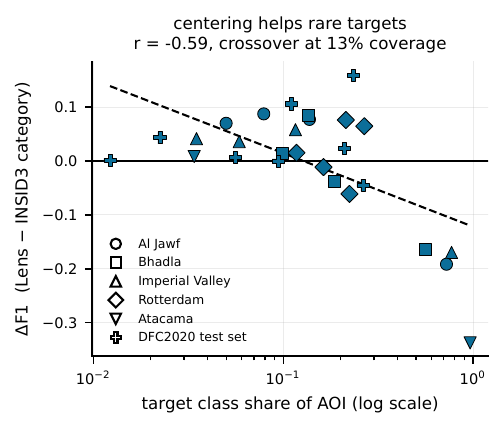}
  \caption{Change in F1 from area centering against target prevalence, over 27 class and corpus combinations. The benefit crosses zero near 0.13.}
  \label{fig:prevalence}
\end{figure}

Centered cosine measures departure from the corpus average. When the target is a small part of the corpus, that is the right thing to measure. When the target is most of the corpus, the average is the target, and there is nothing left to measure. Pooling all 27 class and corpus combinations we evaluated, the Pearson correlation between log prevalence and the change in F1 is $-0.59$, and the fit crosses zero near a prevalence of 0.13 (Figure~\ref{fig:prevalence}). Below that, centering helps in 14 of 17 cases. Above it, it costs as much as 0.34 of F1, the worst case being bare ground at Atacama, which covers 96\% of the area. Site homogeneity, which we had assumed was the controlling variable, is the wrong one: Al Jawf appears on both sides of the line, gaining 0.077 of F1 for cropland at 14\% prevalence and losing 0.192 for bare ground at 72\%.

\subsection{Threshold behavior on a mosaic}

\begin{figure}[t]
  \centering
  \includegraphics[width=\textwidth]{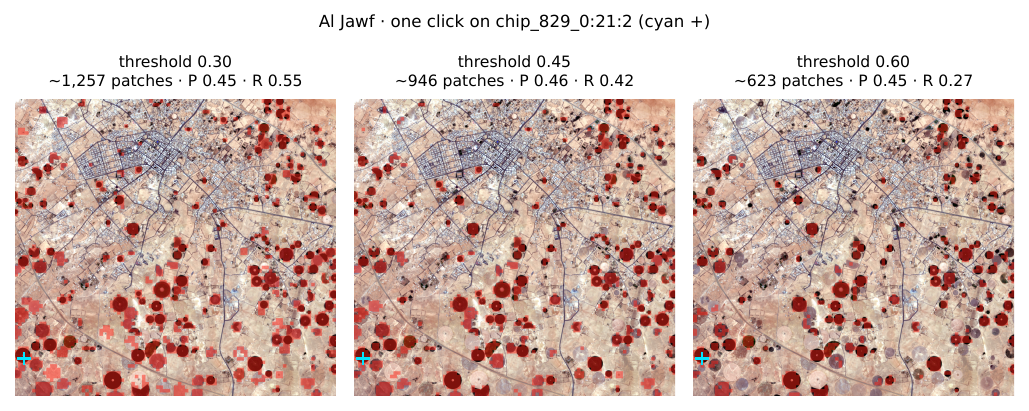}
  \caption{Al Jawf overlays at thresholds 0.30, 0.45, and 0.60 from one click on an irrigation circle. Precision holds at 0.45, 0.46, 0.45 while recall falls from 0.55 to 0.27.}
  \label{fig:thresholds}
\end{figure}

We had expected the slider to move from circles, through vegetated circles, to near-duplicates of the clicked example, and it does not. Precision at Al Jawf is flat across the three thresholds in Figure~\ref{fig:thresholds} while recall falls, so raising the threshold shrinks the returned set, from 1,257 to 623 patch-equivalents, without making it purer. The slider narrows the result predictably rather than refining it semantically, and we withdraw the stronger claim.

\subsection{Scene-level retrieval and the projection at tile level}

We had expected the positional projection to transform tile-level scene retrieval, pulling pairwise pooled cosine out of a degenerate band into a usable range. It does not. Undebiased pairwise cosine already spans a wide range before any correction, from 0.62 to 0.96 at Al Jawf and 0.43 to 0.99 at Atacama, and the projection moves the median by at most 0.012 anywhere. The pooled token holds only 14.5\% of its energy in the positional subspace and holds it almost identically across tiles, so projecting it out acts as a common-mode shift rather than a correction.

The projection earns its place at the patch level, where the index lives. Over 300 DFC2020 tile pairs, a patch's best cross-tile match falls at the same grid position 5.2\% of the time without debiasing, against a chance rate of 0.39\%, so one match in twenty is driven by position rather than content. Debiasing collapses this to 0.8\%, close to chance. That is the failure mode described by \citet{cuttano2026}, and it is severe enough to corrupt a shared index.

\begin{figure}[t]
  \centering
  \includegraphics[width=\textwidth]{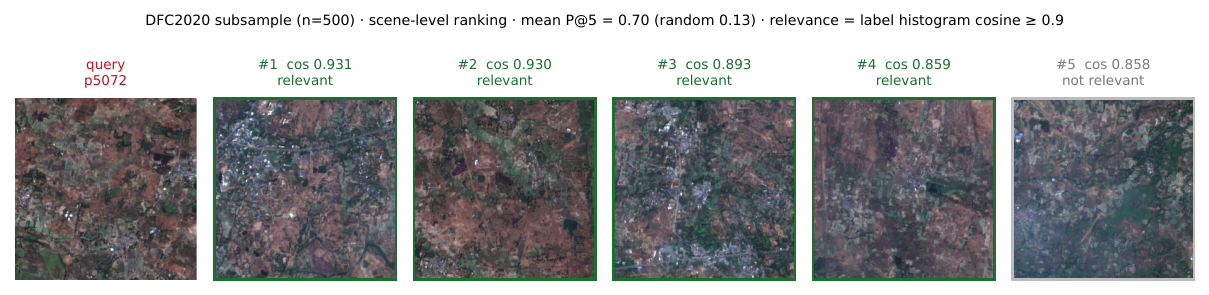}
  \caption{Scene-level retrieval using pooled descriptors. Query tile on the left, top five ranked tiles with cosine values; four of five are relevant.}
  \label{fig:scenes}
\end{figure}

Scene retrieval works regardless. On a fixed subsample of 500 DFC2020 tiles, chosen over the five sites because nine tiles per site makes nearly every tile relevant to every query, ranking by the pooled descriptor reaches precision at 5 of 0.70 and mean average precision at 5 of 0.65 against a random baseline of 0.13, with relevance defined as a label-histogram cosine of at least 0.9 (Figure~\ref{fig:scenes}).

\subsection{Query latency}

Unrefined queries are fast enough that the interface is limited by rendering rather than by search. A click on an area-scale index returns a finished RGBA overlay in 6 to 10~ms at a threshold of 0.30, of which the search itself is 1.7 to 2.7~ms. Lowering the threshold to 0.0 raises the total to 13 to 21~ms, because the returned set grows to several thousand patches. At corpus scale, a range search over 1.31 million patches returns a median of 136,653 hits in 436~ms (Table~\ref{tab:latency}).

\begin{table}[t]
  \centering
  \small
  \caption{Query latency by configuration. Medians on one CPU thread; five-site rows give the range across the areas of Table~\ref{tab:sites}.}
  \label{tab:latency}
  \begin{tabular}{@{}l l l r@{}}
    \toprule
    \textbf{Query type} & \textbf{Corpus} & \textbf{Cache state} & \textbf{Latency} \\
    \midrule
    Click to overlay at 0.30     & Five sites, 9.2k to 12.3k patches & Warm index      & 6.7 to 10.2 ms \\
    Range search at 0.30         & DFC2020, 1.31M patches            & Warm index      & 436 ms \\
    Scene-level top-10           & Five sites                        & Warm index      & 0.014 to 0.02 ms \\
    Refined, first query on area & Five sites                        & Cold            & 26.5 to 29.6 s \\
    Refined, later queries       & Five sites                        & Warm, in memory & 0.20 to 0.21 s \\
    Refined, later queries       & Five sites                        & Warm, from disk & 0.72 to 0.87 s \\
    \bottomrule
  \end{tabular}
\end{table}

Refined queries depend on the cache. The first over a cold area pays about 3~s per tile, and every subsequent refined query on the same area, at any threshold, radius, or click location, takes about 0.2~s from memory, because the work reduces to a dot product against descriptors already computed. This is the payoff from caching descriptors instead of similarity maps, since a cached similarity map would be invalid the moment the analyst moved the click or the slider. Refinement buys a real improvement as well as a visual one: intersection over union rises from 0.325 to 0.370 at Al Jawf and from 0.689 to 0.778 at Bhadla, averaged over six seeds each.

\subsection{Agreement with WorldCover on the five sites}

Table~\ref{tab:worldcover} reports agreement with a land-cover product, not object detection accuracy, and the figures are lower bounds for the object classes. Al Jawf cropland is the clearest example: WorldCover resolves the irrigated area into 4,618 speckled components with a median area of three pixels, against center pivots roughly 500~m across, so a result that correctly outlines whole circles is penalized for covering ground the reference left unmarked. The pattern otherwise matches DFC2020, with built-up and water strongest and vegetation classes distinguished by state rather than structure weakest.

\begin{table}[t]
  \centering
  \small
  \caption{Agreement with ESA WorldCover 2021 v200 \citep{zanaga2022} on the deployed areas, at the per-class best threshold, for classes with sufficient support. Pixel-level F1, means over 30 simulated clicks.}
  \label{tab:worldcover}
  \begin{tabular}{@{}l c c l@{}}
    \toprule
    \textbf{Site} & \textbf{Classes evaluated} & \textbf{F1 range} & \textbf{Strongest class} \\
    \midrule
    Al Jawf         & 4 & 0.28 to 0.66 & Bare / sparse vegetation \\
    Bhadla          & 4 & 0.31 to 0.78 & Built-up \\
    Imperial Valley & 4 & 0.33 to 0.75 & Cropland \\
    Rotterdam       & 5 & 0.37 to 0.78 & Permanent water \\
    Atacama         & 2 & 0.12 to 0.64 & Bare / sparse vegetation \\
    \bottomrule
  \end{tabular}
\end{table}

\subsection{Limitations}

Background bleed persists at low thresholds. Centering removes the shared component but does not equalize variance across feature dimensions, so directions with large residual variance can dominate the cosine. Per-dimension whitening of the centered bank, or subtracting an explicit background prototype, are the obvious next steps and neither requires training.

Scores are relative to the indexed area, since the mean is a property of the area, so a similarity of 0.45 in one area and 0.45 in another do not mean the same thing. Within an area the threshold is consistent, which is what the interface needs. Centering can also hurt, and the system does not yet know when: above a target prevalence of roughly 0.13 the correction costs accuracy, and nothing in the pipeline detects that condition. Compositing discards time as well, since the median over a date window collapses a sequence into one image, which suits stable targets like solar arrays and is wrong for anything that changes within the window. Only three of Sentinel-2's thirteen bands are used, and the DFC2020 results show where this bites: the four weakest classes are the ones a vegetation index would separate.

The evaluation is scoped in four ways. Accuracy is measured against land-cover products rather than instance annotations, so for object classes the scores are lower bounds; we attempted to obtain instance-level center-pivot polygons and could not. The DFC2020 mirror strips georeferencing, forcing us to treat the split as one mixed corpus, the least favorable configuration for centering. All clicks are simulated on reference-positive patches, whereas a real analyst clicks on a clear example and would likely do better. And every timing comes from single-thread CPU execution on one machine. Performing no hand annotation keeps the annotation-free claim intact, but it also means we never had a reference tailored to the objects the system is designed to find.

% ===========================================================================
\section{Conclusion and Recommendation}
\label{sec:conclusion}

Lens shows that a frozen self-supervised vision transformer, with two linear corrections and no training of any kind, supports interactive object and scene retrieval across a satellite mosaic. An analyst clicks one example and receives every similar location, and on DFC2020 that click reaches 0.09 to 0.80 pixel F1 across eight land-cover classes, beating a pairwise baseline on six of them at roughly four hundred times lower query cost.

The two corrections address different problems and do not carry equal weight. Projecting dense descriptors off a positional subspace estimated from a constant image cuts spurious same-position matches from 5.2\% to 0.8\%, which is what allows one index to span many tiles, and it accounts for two thirds of the accuracy the corrections deliver. Subtracting the area mean converts similarity into a measure of departure from the local background, which helps when the target is a small part of that area and hurts when it is most of it, with the crossover near 13\% prevalence. We set out believing centering was the decisive step and report the opposite because that is what the measurements say. The engineering around the corrections decides whether the system is usable: byte quantization holds 1.31 million patches in 515~MB at a similarity error of 0.003, and caching upsampled descriptors rather than similarity maps turns a 28~s wait into a 0.2~s one.

Three directions follow for the system itself. Centering should switch itself off when the returned set covers most of the corpus, and the data needed for that decision is already at hand. Per-dimension whitening of the centered descriptors is the most promising lever against background bleed, and it stays within the training-free constraint. And extending beyond three visible bands is the clearest way to improve the vegetation and wetland classes where we score worst, either through a multispectral encoder or by concatenating the patch descriptors with a published spectral-temporal embedding field. The field would also benefit from a benchmark for point-prompted retrieval with instance-level references, since existing benchmarks score whole-tile ranking and land-cover products penalize a system for outlining objects more completely than the reference does.

Retrieval and detection are complementary rather than competing. A Lens query produces a set of locations an analyst can correct in minutes, and that corrected set is training data for a conventional detector, which is a way to reach the first hundred labels without spending a week.

% ===========================================================================
\section*{Acknowledgements}

Sentinel-2 Level 2A data are provided by the European Space Agency through the Copernicus programme and were accessed through a public SpatioTemporal Asset Catalog endpoint. The DFC2020 tiles come from the 2020 IEEE GRSS Data Fusion Contest \citep{robinson2021} via the Copernicus-Bench mirror, redistributed under CC BY 4.0. Land-cover reference data are from ESA WorldCover 2021 v200 \citep{zanaga2022}. This work uses the DINOv3 models and code released by Meta under the DINOv3 License \citep{simeoni2025}. Built with DINOv3.

% ===========================================================================

\end{document}